\documentclass[10pt,journal,compsoc]{IEEEtran}

\ifCLASSOPTIONcompsoc
  \usepackage[nocompress]{cite}
\else
  \usepackage{cite}
\fi
\ifCLASSINFOpdf
   \usepackage[pdftex]{graphicx}
\else
   \usepackage[dvips]{graphicx}
\fi
\usepackage{amsmath}
\newcommand\degree{{}^\circ}
\DeclareMathOperator*{\argmax}{argmax}

\begin{document}
%
\title{A windowed correlation based feature selection method to improve time series prediction of dengue fever cases}
%
%
%
%


\author{Tanvir~Ferdousi,
        Lee~W.~Cohnstaedt,
        and~Caterina~M.~Scoglio
\IEEEcompsocitemizethanks{\IEEEcompsocthanksitem T. Ferdousi and C. M. Scoglio are with the Department of Electrical and Computer Engineering, Kansas State University, Manhattan,
KS, 66506, USA.\protect\\
E-mail: tanvirf@ksu.edu
\IEEEcompsocthanksitem L. W. Cohnstaedt is with the Arthropod-Borne Animal Diseases Research Unit, Center for Grain and Animal Health Research, USDA, Manhattan, KS, 66502, USA.}
\thanks{Mention of trade names or commercial products in this publication is solely for the purpose of providing specific information and does not imply recommendation or endorsement by the U.S. Department of Agriculture. The conclusions in this report are those of the authors and do not necessarily represent the views of the USDA. USDA is an equal opportunity provider and employer.}}

\IEEEtitleabstractindextext{%
\begin{abstract}
The performance of data-driven prediction models depends on the availability of data samples for model training. A model that learns about dengue fever incidence in a population uses historical data from that corresponding location. Poor performance in prediction can result in places with inadequate data. This work aims to enhance temporally limited dengue case data by methodological addition of epidemically relevant data from nearby locations as predictors (features). A novel framework is presented for windowing incidence data and computing time-shifted correlation-based metrics to quantify feature relevance. The framework ranks incidence data of adjacent locations around a target location by combining the correlation metric with two other metrics: spatial distance and local prevalence. Recurrent neural network-based prediction models achieve up to 33.6\% accuracy improvement on average using the proposed method compared to using training data from the target location only. These models achieved mean absolute error (MAE) values as low as 0.128 on [0,1] normalized incidence data for a municipality with the highest dengue prevalence in Brazil's Espirito Santo. When predicting cases aggregated over geographical ecoregions, the models achieved accuracy improvements up to 16.5\%, using only 6.5\% of incidence data from ranked feature sets. The paper also includes two techniques for windowing time series data: fixed-sized windows and outbreak detection windows. Both of these techniques perform comparably, while the window detection method uses less data for computations. The framework presented in this paper is application-independent, and it could improve the performances of prediction models where data from spatially adjacent locations are available.
\end{abstract}

\begin{IEEEkeywords}
Dengue, Feature Selection, Machine Learning, Recurrent Neural Networks, Time Series.
\end{IEEEkeywords}}


\maketitle

\IEEEdisplaynontitleabstractindextext

%
\IEEEpeerreviewmaketitle

\ifCLASSOPTIONcompsoc
\IEEEraisesectionheading{\section{Introduction}\label{sec:introduction}}
\else
\section{Introduction}
\label{sec:introduction}
\fi

%
%
%
%
\IEEEPARstart{A}{ccurate} time series prediction of dengue fever outbreaks can be useful in planning mitigation strategies for hundreds of tropical and subtropical regions around the world. Data-driven models such as neural networks are flexible in design and can ease the difficulties of estimating unknown parameters that mechanistic models often require \cite{baker2018mechanistic}. However, the prediction accuracy of such models depends on the quality and the quantity of training data. For dengue fever outbreaks, the availability of incidence data varies across regions. A data aggregation center in a region may not have adequate data to achieve an acceptable level of accuracy in out-of-sample projections. In such cases, selected incidence data from adjacent centers in the same region could improve model performance as additional features. We propose a framework with quantitative methodologies to rank and select nearby case data as supplementary features. Our method uses windowed time-lagged cross-correlation combined with distance and prevalence metrics to identify relevances and potential causal relationships.

Dengue virus is primarily spread by several species of mosquito vectors (\textit{Aedes aegypti} and \textit{Aedes albopictus}), and the outbreaks infect 390 million people every year \cite{bhatt2013global}. The viral strains also cause about 40,000 annual deaths with hemorrhagic fever, and dengue shock syndrome \cite{roth2018global}. Dengue virus transmission is prevalent in regions where competent vector mosquitoes are present. In those regions, the mosquito abundance varies throughout the year and depends on factors including air temperature, precipitation, vegetation, and urbanization \cite{li2019climate, fuller2009nino, wu2009higher}. The availability of extensive data on these factors makes statistical and machine learning analyses feasible. However, researchers experience missing data, uneven reporting intervals, lack of granularity, inadequacy, and inaccuracy with dengue case counts for many regions around the globe. These factors limit the prediction performance of data-driven models. To complicate the situation further, time series outbreak data for most diseases are non-stationary (e.g., the underlying processes evolve with time). In many regions, climatic variations affect \textit{Aedes} vector populations \cite{wongkoon2013distribution}. In addition to that, outbreak start times and sizes may vary because of imported cases caused by short and long-range mobility. Co-circulation of multiple viral strains adds to the complexity. Despite those issues, correlations exist between outbreaks in adjacent human populations (county, municipality, district, etc.) for most communicable diseases, including dengue \cite{mammen2008spatial}. While a correlation may not always imply causation, the use of incidence data from adjacent regions can improve the model training because of similarities in meteorological factors, host population density, and a high probability of mobility-based viral transmissions.

For sequential data (e.g., multivariate time series), different methods of feature selection have been used in the past including correlation-based filters \cite{gonzalez2016towards, moreno2016applicability}, Granger causality tests \cite{sun2015using, hmamouche2017causality}, genetic algorithm \cite{garrett2003comparison}, and several other methods \cite{crone2010feature, munkhdalai2019end}. The applications include forecasts of electrical energy consumption, meteorological variables, financial markets, etc. Correlation-based methods are widely used in machine learning problems \cite{yu2003feature, hall1999correlation}. Few works extend feature sets for disease outbreak prediction using incidence data from spatially adjacent locations. Such geospatial clustering techniques rely on similarity metrics to improve model performance. A recent work \cite{liu2020real} uses pair-wise correlation to cluster location data to train models for COVID-19 outbreak projection in Chinese provinces. Another work targeted towards dengue also uses correlation-based similarity measures to extend the training feature set \cite{mussumeci2020large}. However, these implementations assume an instantaneous correlation of incidence data between regions and do not consider the temporal order of outbreaks (whether one location is leading or lagging the other). There can be a considerable amount of time delay between outbreaks occurring in adjacent regions. Time lagged cross-correlation can help identify such phase relations \cite{schoenherr2019identification}. The phase information may help quantify relationships between outbreaks of adjacent locations. We hypothesize that \textit{if the temporal incidence dynamics of one outbreak lead the incidence dynamics of another, the former outbreak might have a causal influence on the latter}, given that these places lie spatially close enough to affect one another, and the outbreaks are of a reasonable size. We weigh the available features (incidence data) based on a combination of these factors: leading phase correlation, distance, and prevalence. Because of the population's dynamically changing immunity patterns caused by the co-circulation of multiple dengue virus strains, the regions experiencing large outbreaks may also evolve. A single computation of the correlation coefficient over the entire timeline of data may mislead the analysis. Hence, splitting the time series into multiple windows and comparing sequences at each time window for correlation and phase analyses can provide better insights into the dengue outbreak's seasonal patterns. A limitation with some existing methods is that clustering large data sets for model training can make the process complex and inefficient, while unrelated features may deteriorate prediction performance. This phenomenon is known as the curse of dimensionality \cite{hastie2009elements,li2017feature}. This work aims to reach optimal model performance with the smallest subset of highly relevant features based on their ranks. There are two broad categories of feature selection methods: filters and wrappers \cite{saeys2007review}. Wrapper approaches \cite{jimenez2020feature} are computationally expensive as the search space for optimal feature subsets increases exponentially with the number of available features. Our method is primarily a filter approach to rank features.

In this work, we present a framework of feature enhancement for data-driven prediction of dengue outbreaks. To achieve that goal, this paper details: i) a windowed time-shifted cross-correlation method to compute correlation weights, ii) two correlation-window allocation methods, iii) a procedure of ranking feature using metrics based on correlation, distance, and prevalence, and iv) analysis of prediction performance across windowing schemes, prediction models, and spatial aggregations. This work's novel contribution lies in how we interpret and process incidence data to compute correlation metrics using our knowledge of how outbreaks spread.

\section{Preliminaries}
\label{sec:preliminaries}
\subsection{Definitions}
In supervised learning problems, we collect data on multiple variables. A \textit{target} variable (\textit{label}) is the designated output of a machine learning model for prediction. A \textit{feature} is an input variable that is expected to influence the target variable. Each \textit{instance} of data (e.g., a point in time) contains several feature values and usually a single label value. A data set comprises many such instances. For a supervised learning problem, a data set is split into 3 subsets in order to: \textit{train}, \textit{evaluate}, and \textit{test} the models. With the training subset fed in batches (collection of instances), a supervised model learns to predict targets based on features in an iterative process. It optimizes parameters by minimizing a loss function. A neural network comprises artificial neurons (cells) in one or multiple layers. Each neural cell is a node in the network with connections to other nodes across layers, inputs, or outputs. The recurrent neural networks (RNN) are special neural cells that can store internal states in their memory, which helps them predict sequence data (data points that are temporally related) better. Model training aims to get optimal parameter values to generalize beyond the training data and perform well with test data. Sometimes, a model can over-fit the training data and perform poorly with unseen test examples. To prevent such scenarios, we use regularization techniques.

\subsection{Time series prediction of outbreaks}
Time series forecasting is a popular research area because of its applicability in many disciplines, such as forecasting weather patterns, stock prices, market trends, and resource allocation. In epidemiology, these methods enable the prediction of future outbreaks by fitting models with past disease incidence data and carefully chosen covariates. Such predictions come at varying degrees of accuracy and depend on the characteristics of the target variable, quality of sample data available for fitting, the covariates (predictors) being used, and the models themselves. There is rarely a single model that works best for every application. Classical forecasting methods such as exponential smoothing, autoregressive integrated moving average (ARIMA), and seasonal autoregressive integrated moving average (seasonal ARIMA) have been widely used to predict time series data \cite{box2011time}. The ARIMA model can handle non-stationary data, which is an important advantage. Besides that, the seasonal ARIMA model can incorporate repeating patterns in the data to enable forecasting of diseases that show seasonal patterns \cite{permanasari2013sarima}. However, these models have tendencies to follow the mean values of past data, and it is not easy to associate these with rapidly changing processes \cite{hong2012application}. In addition to that, many classical models require manual tuning of their parameters and may fail to capture complex nonlinear interactions. Data-driven forecasting of vector-borne diseases such as dengue fever is complicated due to complex interactions of several factors with disease dynamics. A list of these factors include but is not limited to seasonally dependent \textit{Aedes} mosquito growth and feeding patterns \cite{scott1993blood}, co-circulations of multiple viral strains \cite{shrivastava2018co}, environmental (e.g., temperature) effects on dengue virus transmission \cite{huber2018seasonal}, and human mobility patterns\cite{wesolowski2015impact}. Neural networks can automatically interpret features from observable variables and can model complex nonlinear phenomena. Hybrid methodologies that combine neural networks with classical models (e.g., ARIMA) are also popular and have been used for dengue outbreak forecasting \cite{chakraborty2019forecasting}. In recent times, long-short term memory (LSTM) networks \cite{hochreiter1997long} (a type of recurrent neural networks), and its derivatives \cite{laptev2017time, hua2019deep} have shown a considerable amount of success in predicting sequential data and has frequently outperformed other classical and machine learning methods \cite{zhu2017deep}. These recurrent architectures have also performed well to predict disease outbreaks \cite{gu2019method, shahid2020predictions, xu2020forecast}. Hence, we consider these viable candidates to test the performance of the proposed feature enhancement framework in this paper.

\subsection{Factors that affect dengue disease dynamics}
Dengue virus is primarily spread by female mosquito vector species: (\textit{Aedes aegypti} and \textit{Aedes albopictus}). Hence, the outbreaks depend on the abundance of such vectors. The relationships between \textit{Aedes} mosquitoes and environmental variables (i.e., temperature, rainfall) are already well characterized by many researchers. Environmental temperature affects the growth, host-seeking, blood-meal intakes of mosquitoes. \textit{Aedes aegypti} cannot develop below $16\degree$ C or above $34\degree$ C \cite{christophers1960aedes}. Within that range, the development from larva to adult was found to be faster at higher temperatures ($30\degree$ C) compared to lower temperatures ($21\degree$ C) \cite{couret2014temperature}. \textit{Aedes albopictus} can develop in wider temperature ranges and can survive better in lower temperatures \cite{delatte2009influence} compared to \textit{Aedes aegypti}. The optimum flight temperature for \textit{Aedes aegypti} females was found to be $21\degree$ C \cite{rowley1968effect}. Studies have observed that a large diurnal temperature range decreases female fecundity \cite{carrington2013large}. Temperature fluctuations also affect extrinsic incubation periods (EIP) of dengue viruses. An experiment with DEN-2 strain found that EIP was 12 days at $30\degree$ C and reduced to 7 days for $32\degree$ C and $35\degree$ C \cite{watts1987effect}. Besides temperature, rainfall has a significant role in dengue outbreaks. Rainwater stuck in different places creates breeding spaces for \textit{Aedes} mosquitoes. Previous works have studied the association of dengue transmission with rainfall \cite{johansson2009local, li2019climate}. In this work, we use several variables, including observed and reanalyzed temperatures, precipitations, relative humidity, and surface-level pressure. These can directly or indirectly affect mosquito vector suitability and dengue outbreak dynamics.

\subsection{Data collection and processing}
\subsubsection{Data acquisition}
To test the proposed framework for dengue outbreak predictions, we collect data for several regions of Brazil. The InfoDengue project \cite{infoDengue} monitors outbreak data on over 700 municipalities of Brazil. Their server contains weekly dengue fever case counts with a surveillance period starting from 2010. Besides dengue incidence data, we collect weather observation data from NOAA (National Oceanic and Atmospheric Administration) ground weather station database and reanalysis data from the NCEP /NCAR Reanalysis 1 dataset published by the NOAA physical sciences laboratory (PSL) database (NCEP and NCAR stand for National Centers for Environmental Prediction and National Center for Atmospheric Research, respectively). We list all the variables in Table \ref{table:dataSources}. We use the weekly case counts as labels and further process the remaining variables to use those as baseline features.

\begin{table}[ht]
\caption{Data collection sources}
\label{table:dataSources}
\centering
\begin{tabular}{|p{2.5cm}|p{.8cm}|p{2cm}|p{2cm}|}
\hline
Variable Name & Time Resolution & Space Resolution & Source \\
\hline\hline
Reported dengue fever cases & Weekly & Municipality level & INFO Dengue \cite{infoDengue} \\
Observed Temperatures (min, max, and avg) & Daily & Ground station dependent & NCEI-NOAA \\
Observed Precipitation & Daily & Ground station dependent & NCEI-NOAA \\
Avg surface air temperature & Daily & $2.5 \degree \times 2.5 \degree$ & NCEP/NCAR Reanalysis 1 \\
Avg surface relative humidity & Daily & $2.5 \degree \times 2.5 \degree$ & NCEP/NCAR Reanalysis 1 \\
Avg surface pressure & Daily & $2.5 \degree \times 2.5 \degree$ & NCEP/NCAR Reanalysis 1 \\
Avg precipitable water & Daily & $2.5 \degree \times 2.5 \degree$ & NCEP/NCAR Reanalysis 1 \\

\hline
\end{tabular}
\end{table}

\subsubsection{Re-sampling and feature engineering} \label{sec:feat_engg}
The available raw data cannot be readily used in the machine learning methods. All the features and labels are matched and aligned in both spatial and temporal dimensions. We align data based on available labels (i.e., case data). For each municipality of Brazil (smallest spatial unit available), we search for the nearest ground weather station to collect weather data. We extract reanalysis data from NCEP/NCAR Reanalysis 1 data sets using each municipality's centroid's coordinates. Once the spatial granularity is taken care of, we fix the temporal dimension mismatches by converting all data to match incidence data resolution. As incidence data are available in weekly intervals and the remaining variables are available daily, this process involves context-aware down-sampling of those remaining variables (temperature, precipitation, humidity, etc.). During this process, we derive 4 additional features from observed weather data of ground stations: average diurnal temperature range of the week, minimum diurnal temperature range of the week, maximum diurnal temperature range of the week, and the number of rainy days in the week. We compute these from daily observed temperature and precipitation data. In total, we have 12 feature variables related to weather and environment: i) 4 observed variables: temperature (average, minimum, and maximum) and precipitation, ii) 4 derived variables based on the time interval (week): diurnal temperature range (average, minimum, and maximum) and the number of rainy days, iii) 4 reanalysis variables: temperature, relative humidity, pressure, precipitable water (all are averages and at earth surface level).

\subsubsection{Data splitting and normalization}
The complete set of data is a two-dimensional array $X$ with \textit{features} on one dimension and \textit{time} on the other. The set of features consists of: i) the variables listed in section \ref{sec:feat_engg}, ii) dengue incidence data of the target location, and iii) dengue incidence data of locations selected as predictors by the methods presented in this paper. We split $X$ along its time dimension into three parts: training ($X_{train}$), validation ($X_{val}$), and test ($X_{test}$). The validation set is required during the training phase as we implement \textit{early stopping}\cite{prechelt1998early} as a regularization mechanism. We normalize all three sets of data before model training and evaluation. The normalization formula is the following,

\begin{align}
    \mu_{train} &= MEAN(X_{train}) \nonumber \\
    \sigma_{train} &= SD(X_{train}) \nonumber \\
    \hat{X_{train}} &= (X_{train} - \mu_{train})/\sigma_{train} \nonumber \\
    \hat{X_{val}} &= (X_{val} - \mu_{train})/\sigma_{train} \nonumber \\
    \hat{X_{test}} &= (X_{test} - \mu_{train})/\sigma_{train}
    \label{eqn:data_norm}
\end{align}

All three data sets (training, validation, and test) are normalized using the mean and the standard deviation computed from training data. The complete data set's summary statistics are not used here to prevent the machine learning models from gaining statistical insights about validation and test data sets.

\subsection{Sequence model specifications}
\label{sec:model_spec}
\begin{figure}[ht]
\centering
\includegraphics[width=.8\linewidth]{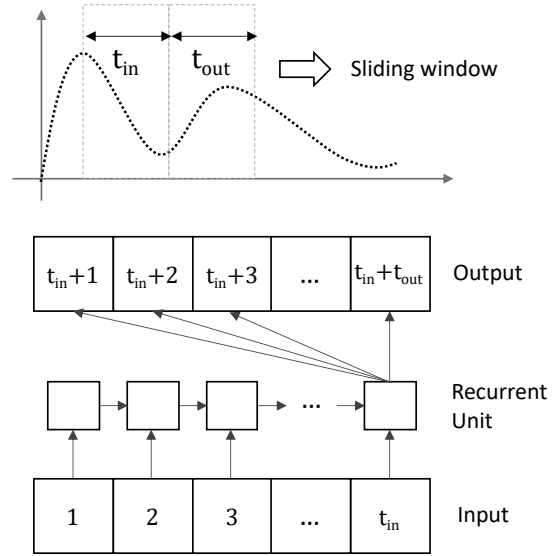}
\caption{A sliding window defined for feeding of input data and extraction of output case counts. The window slides along the horizontal axis and feeds $t_{in}$ time steps of feature data into the model and extracts $t_{out}$ time steps of predictions.}
\label{fig:sliding_window_rnn}
\end{figure}

Whether it is training or evaluation, the time series data are split into smaller batches and fed into the recurrent neural network models. From a macroscopic perspective, a sliding window moves over batches of data. Because of the sequential nature of dengue case data, the batches are fed according to the time order without randomization. The sliding window is configured with two integer parameters: input length ($t_{in}$), and output length ($t_{out}$). The window is depicted in Figure \ref{fig:sliding_window_rnn}. A trainable model would take $t_{in}$ time steps of feature data as input and predict $t_{out}$ time steps of target/label data (e.g., dengue case counts) as outputs every iteration. In the configurations used in this paper, there is no temporal overlap between input and output sequences. In this configuration, the models make single shot projections (all the $t_{out}$ data points are predicted at once every iteration).

Each batch (window) of data is ($t_{in}+t_{out}$) steps long in the time dimension. A total of 32 batches are stacked together for model training and evaluation. One batch differs from another by a single time-step (hence, there are temporal overlaps between batches). Each batch is further split in time and data (variable) dimensions to separate inputs ($t_{in}$ of features) and outputs ($t_{out}$ of labels).

This work focuses on performance gains obtainable using recurrent neural network (RNN) models due to their proven track record in predicting time series data \cite{zhu2017deep}. A recurrent unit's temporal behavior is illustrated in the lower part of Figure \ref{fig:sliding_window_rnn}. We can see that information is passed through time (also called cell state), enabling the model to predict values based on insights gained from past inputs. We use two popular recurrent neural network cell types: LSTM (long-short term memory)\cite{hochreiter1997long} with forget gates \cite{gers2000learning} and GRU (gated recurrent unit)\cite{cho2014learning}. We also test with a simple linear neural network model as a trivial baseline. Using the TensorFlow\cite{abadi2016tensorflow} package, we configure the models as described below to produce the results:

\begin{itemize}
  \item \textbf{Linear}: The model consists of a single layer of artificial neurons (\textit{Dense} units in TensorFlow) without any nonlinear activation functions. The size of the layer (number of neural units) is equal to the output prediction steps, $t_{out}$.
  \item \textbf{LSTM}: The model consists of an input layer of 32 long short-term memory (LSTM) units. The output layer consists of a layer similar to the Linear model (described above).
  \item \textbf{GRU}: The model consists of an input layer of 32 gated recurrent units (GRU). The output layer consists of a layer similar to the Linear model (described above).
\end{itemize}

We initialize the weight metrics of the models as zeros in the beginning. Only the recurrent models (LSTM and GRU) can train and predict based on entire input sequences. The Linear model predicts based on the last input time step. While training, we use the mean squared error (MSE) as the loss function to optimize the model using Adam optimizer \cite{kingma2014adam}. For predictions, we use the mean absolute error (MAE) metric to evaluate model performance. We regularize our training process with the early stopping\cite{prechelt1998early} mechanism, which monitors loss within validation data and stops training if performance does not improve. Based on our tests on different data sets, we configure the training to run for 120 iterations (epochs).

\section{Methods}
\label{sec:methods}
\begin{figure}[ht]
\centering
\includegraphics[width=\linewidth]{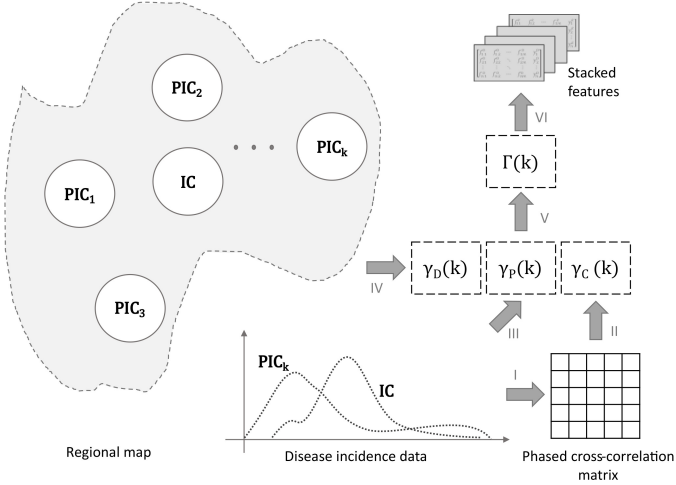}
\caption{A framework depicting the method to expand trainable and testable data set on dengue incidence. The shaded region enclosed by dashed borders on the left depicts a map of the region of interest. The circles inside the map indicate multiple infection centers ($IC$) across the region. The target infection center is marked as $IC$, while the $k^{th}$ peripheral infection center is $PIC_k$. Using the proposed windowed cross-correlation method (section \ref{sec:win_corr_method}), a phased cross-correlation matrix is computed (I), which is eventually reduced to a correlation weight, $\gamma_C (k)$ (II). We also compute a prevalence weight, $\gamma_P (k)$, using cumulative case data (III) and a geographic distance weight, $\gamma_D (k)$, using location data (IV). The three weight metrics are combined to compute (V) the predictor metric of $PIC_k$, $\Gamma(k)$. The $PIC$s are ranked using these $\Gamma$ values, and their incidence data are selected accordingly to be stacked together with the $IC$ feature set (VI).}
\label{fig:framework}
\end{figure}

To describe the methodology, first, we define our spatial units. We designate the term \textit{infection center} ($IC$) to indicate a spatial building block of the model. An $IC$ is a point in the space (regional map) where observed or estimated incidence data on disease outbreaks are available. The spatial granularity of an $IC$ is not fixed for the model. It can be a country, a state, a city, a suburb, or an administrative unit with some resolution in space based on available disease incidence data. The basic structure of our proposed framework is shown in Figure \ref{fig:framework}. The circles inside the shaded region are the infection centers. One of the infection centers, marked as $IC$, indicates the target infection center where we intend to make predictions of a designated label (e.g., dengue cases). The map's remaining infection centers are marked as \textit{peripheral infection centers} ($PIC$). These are locations where similar observations on the designated label of the target $IC$ are available. The temporal resolution of the $IC$ and the $PIC$s are aligned before performing any comparative analysis of the data. Some locations may have daily, weekly, or monthly observations. Some data may need re-sampling in the time domain before they can be compared (e.g., convert daily weather data to weekly values).

Assuming that there are $N$ peripheral infection centers ($PIC$) on the map. Once we match the spatial and the temporal dimensions of the label data (e.g., weekly dengue cases in a city), we use a windowed-time shifted cross-correlation analysis on each $IC-PIC_k$ pair (for all $k \in N$) and compute a correlation weight, $\gamma_C (k)$. We also consider the cumulative cases of each $PIC$ and compute a prevalence weight, $\gamma_P (k)$. Finally, the geodesic distance of each $IC-PIC_k$ pair is taken into account in the form of a distance metric, $\gamma_D (k)$. All three metrics are normalized and lie within the range $[0,1]$ for the selected region. These are combined as following to compute a predictor strength metric for each $PIC_k$,

\begin{equation}
\Gamma(k) = \gamma_C (k)[\gamma_P (k) + \gamma_D (k)], \forall k \in N
\label{eqn:pred_strength}
\end{equation}

\subsection{Windowing incidence data}
\label{sec:win_data}
\begin{figure}[ht]
\centering
\includegraphics[width=\linewidth]{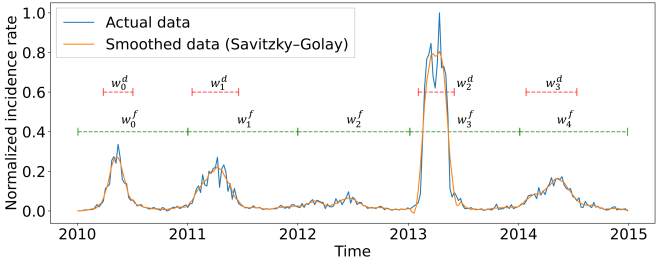}
\caption{Windowing methods used for data segmentation before computing time-shifted correlation coefficients. We present the normalized incidence data from the \textit{Cachoeiro de Itapemirim} municipality of Brazil \cite{infoDengue}. The fixed windows ($w^f_m$) are marked in green, and the detected windows ($w^d_m$) are marked in red. We set $M^f = 5$ to get 5 fixed windows, each comprising 1 year of data. For the detected windows, we configure $\Delta_{MIN} = 10$ and $i_{MIN} = 0.05$. A Savitzky-Golay\cite{savitzky1964smoothing} smoothing is applied to the data before window detection takes place.}
\label{fig:win_methods}
\end{figure}

We compare the time series of disease incidence data (e.g., weekly cases per 100k people) to find correlations. The comparisons are made for all $IC-PIC_k$ pairs with available data for a given region. The key intuition behind this approach is, if a $PIC$ in the region has an infectious outbreak at some point in time, $t = t_0$, this may initiate or affect the course of an outbreak for the target $IC$ at $t \geq t_0 $. A leading $PIC$ outbreak may not always imply causal influence depending on the geographic location and population behavior. Despite that, a $PIC$ having an outbreak will influence adjacent $ICs$, as it acts as an infection source. This also assumes that a strict isolation measure is not in place and the control measures are not 100\% effective due to the vector-borne nature of the infection. It is also important to note that, despite seasonal patterns, outbreaks can occur irregularly. A $PIC$ may lead the target $IC$ in one season and lag in other seasons due to complex interactions of multiple viral strains. Hence, we divide the time series into smaller time windows. We propose two methods for windowing incidence data: i) fixed-length window allocation and ii) variable-length window detection. Both of these methods are depicted in Figure \ref{fig:win_methods}.

A straightforward approach is to divide the entire time series into a fixed number of intervals, $M^f$. If the time series is $T$ units (day, week, or month) long in total, then fixed window, $w^f_m$ (where $m \in [0, M^f-1]$) will have $T/M^f$ units of data. While this is the simplest way to divide the series for correlation analysis, it may not be the most efficient. Setting the appropriate value of $M_f$ remains an open problem, although we apply some intuitions from the seasonality patterns of dengue outbreaks in our case. The fixed windows might not be appropriately placed to contain the time series curve's meaningful dynamics, and some windows may even cover regions without an outbreak.

A second approach is to detect windows based on the time series itself. To do this, we normalize the incidence rate of the target $IC$ to be ranged between $[0,1]$. A typical outbreak curve has irregularities in its shape that make the analysis cumbersome. We use a Savitzky-Golay filter \cite{savitzky1964smoothing} to smooth out the irregularities while preserving the shape of the outbreaks. Our window detection method has two parameters: the incidence threshold ($i_{MIN}$) and the minimum window size ($\Delta_{MIN}$). An window is detected between two time points, $t_{START}$ and $t_{END}$ (where, $0 \leq t_{START} \leq t_{END} \leq T$), if the normalized incidence rate, $i_N(t) > i_{MIN}$ for all $t_{START} \leq t \leq t_{END}$ and $t_{END} - t_{START} \geq \Delta_{MIN}$. A detected window, $w^d_m$ will have a length (greater than $\Delta_{MIN}$) that depends on the time series curve characteristics. The number of detected windows, $M^d$, will also vary for the same reason. Figure \ref{fig:win_methods} shows both methods in action using time series data for Brazil's municipality between 2010-2015. For the assigned value of $M^f=5$, we get equally sized windows, each of which is 1 year in length. With our detection method, we find 4 windows ($M^d = 4$) that indicate 4 separate outbreaks ($\Delta_{MIN} = 10, i_{MIN} = 0.05$).

Once the windows are selected (either by assignment of fixed number or detection), the following procedures are identical. Hence, we will ignore the superscripts ($f$ and $d$) in this paper's next sections for brevity. $M$ will depict the total number of windows. $w_m$ (where $m \in [0, M-1]$) will depict the $(m+1)^{th}$ window.

\subsection{Time-shifted cross correlation}
\label{sec:win_corr_method}

Let $i_0(t)$ and $i_k(t)$ be the disease incidence (new cases at time step $t$) of the target $IC$ and the $k^{th}$ $PIC$ respectively. We perform bivariate computations of time shifted Pearson's correlation coefficients \cite{benesty2009pearson} using windowed ($w_m$) incidence data of the target $IC$ ($i_{0}^{w_m} (t)$) and the $k^{th}$ $PIC$ ($i_{k}^{w_m} (t)$). The formula used to compute the coefficients is shown in Equation \ref{eqn:phased_corr}. This measure is also known as time lagged (or phased) cross-correlation \cite{xCorr} in statistics and signal processing.

 \begin{equation}
 R_{k}^{w_m} (\theta) = r(i_{0}^{w_m} (t), i_{k}^{w_m} (t-\theta))
    \label{eqn:phased_corr}
 \end{equation}
 
Here, $R_{k}^{w_m} (\theta)$ is the correlation coefficient computed for subsets of the time series $i_0(t)$ and $i_k(t)$ selected by the time window $w_m$ when one series is shifted by an amount $\theta$ with respect to another. The Pearson's correlation coefficient function is indicated by $r()$ in Equation \ref{eqn:phased_corr}.

\begin{figure}[ht]
\centering
\includegraphics[width=\linewidth]{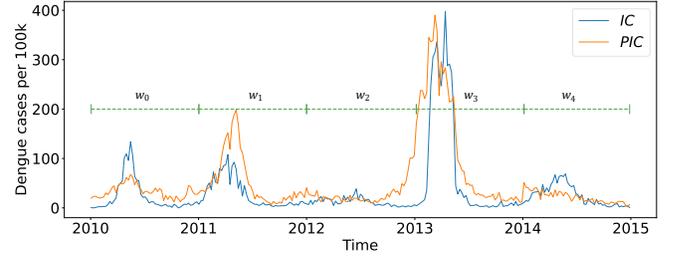}
\caption{Windowing of $IC$ and $PIC_k$ incidence data for computing time-shifted cross-correlation coefficients. We allocate a fixed number of windows ($M^f=M=5$) for the time range 2010-2015, making each window 1 year long (52 weeks approximately). The two curves shown here correspond to the target $IC$ ($i_0(t)$) and the $k^{th}$ $PIC$ ($i_k(t)$) and these are from the municipalities: \textit{Cachoeiro de Itapemirim} and \textit{Vitória} respectively \cite{infoDengue}. The $m^{th}$ time window is marked by the symbol $w_m$. Note, in the first window ($w_0$), the outbreaks of $IC$ and $PIC$ are almost in phase, whereas in the second window ($w_1$), the $PIC$ outbreak is lagging in time compared to the $IC$.}
\label{fig:corr_window}
\end{figure}

\begin{figure}[ht]
\centering
\includegraphics[width=\linewidth]{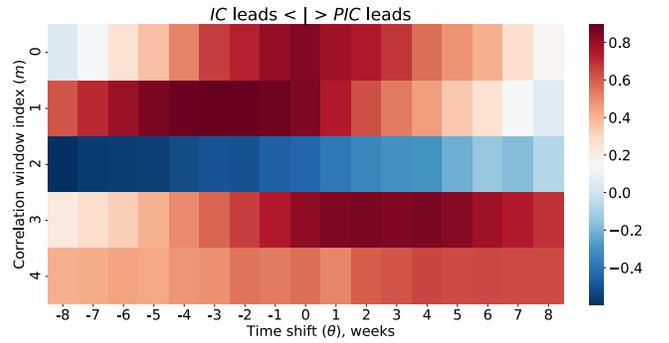}
\caption{Heatmap of the computed time-shifted cross-correlation matrix (Equation \ref{eqn:phased_corr}) for the $IC$ and the $PIC$ in Figure \ref{fig:corr_window}). The vertical axis depicts the window indices ($m \in M$) and the horizontal axis depicts time shift (phase), $\theta$ that ranges from $-8$ to $+8$ weeks. The color shades of the heatmap depict the correlation values demonstrated by the gradient bar on the right. Higher correlation values on the left of the midpoint ($\theta < 0$) indicate that $IC$ is leading in the outbreak curve (Figure \ref{fig:corr_window}) compared to the $PIC$. Higher correlation values on the right of the midpoint ($\theta > 0$) indicate the opposite ($PIC$ leads $IC$).}
\label{fig:corr_heatmap}
\end{figure}

For the two time series curves shown in Figure \ref{fig:corr_window}, the time-lagged correlation matrix (obtained by computing $R_{k}^{w_m} (\theta)$ $\forall$ $m \in [0, M-1]$ and $\theta \in [-8, 8]$) is plotted as a heatmap in Figure \ref{fig:corr_heatmap}. We use the location depicted in Figure \ref{fig:win_methods} as the target $IC$ ($i_0(t)$) and another location from the same state (Espirito Santo) of Brazil as the $k^{th}$ $PIC$ ($i_k(t)$). For this demonstration, the time series curves were split using fixed length windows ($M^f=M=5$). The heatmap depicted in Figure \ref{fig:corr_heatmap} can be visually verified by comparing with Figure \ref{fig:corr_window}. As expected, the two curves are almost in phase in $w_0$ , negatively correlated in $w_2$, $IC$ leads in $w_1$, and $PIC$ leads in $w_3$ and $w_4$.

Heatmaps like Figure \ref{fig:corr_heatmap} are computed for all $PIC_k$ with $k \in [1,N]$. To determine if a $PIC_k$ is leading in a window ($w_m$), we find the location ($\theta$) of the peak correlation as shown in Equation \ref{eqn:theta_peak}.

 \begin{equation}
 \theta_{k}^{w_m} = \argmax_\theta R_{k}^{w_m} (\theta)
    \label{eqn:theta_peak}
 \end{equation}
 
We define the correlation strength $S_k^{w_m}$ to be the mean correlation measure computed around the peak ($\theta_{k}^{w_m}$), extending by the amount $\theta_E$ in both directions (Equation \ref{eqn:mean_corr_strength}).

\begin{equation}
 S_k^{w_m} = \frac{1}{\sum_{\theta= -\theta_E}^{\theta_E} 1} \sum_{\theta= -\theta_E}^{\theta_E} R_{k}^{w_m} (\theta_{k}^{w_m}+\theta)
    \label{eqn:mean_corr_strength}
 \end{equation}

Equation \ref{eqn:mean_corr_strength} has an additional condition on the values $\theta$ such that $R_{k}^{w_m}(\theta_{k}^{w_m}+\theta)$ exists for the given parameters. A $PIC_k$ leads the $IC$ if the peak of correlation lies on the right half of the heatmap shown in Figure \ref{fig:corr_heatmap}, which translates to $\theta_{k}^{w_m} > 0$. We only consider if a $PIC_k$ is at least in phase with the $IC$ and discard the cases where any $PIC_k$ lags the $IC$. This is how we compute the predictor probability matrix $P$ with dimensions $M \times N$. The individual predictor probabilities ($\forall m \in [0,M-1]$, $\forall k \in [1,N]$) are are computed as shown in Equation \ref{eqn:pred_prob_mat}.

\begin{equation}
P_{m,k}= 
\begin{cases}
    S_k^{w_m},& \text{if } \theta_{k}^{w_m}\geq 0\\
    0,        & \text{otherwise}
\end{cases}
\label{eqn:pred_prob_mat}
 \end{equation}
 
The predictor probabilities are averaged across all windows (Equation \ref{eqn:mean_pred_prob}) to compute overall predictive abilities of all $PIC_k$. For a particular region ($k\in [1,N]$), the predictive ability metrics are normalized between [0,1]. The final measure, $\gamma_C(k)$, is defined as the \textit{correlation weight} of $PIC_k$ as shown in Equation \ref{eqn:corr_weight}.
 
\begin{equation}
    \hat{\gamma_C(k)} = \frac{1}{M} \sum_{m=0}^{M-1} P_{m,k}
    \label{eqn:mean_pred_prob}
\end{equation}
\begin{equation}
    \gamma_C(k) = \frac{\hat{\gamma_C(k)} - \min_k \hat{\gamma_C(k)}}{\max_k \hat{\gamma_C(k)} - \min_k \hat{\gamma_C(k)}}
    \label{eqn:corr_weight}
\end{equation}

\subsection{Distance and prevalence metrics}
With increasing distance, the impact of a $PIC$ on the target $IC$ is likely to be reduced due to decreased travel between the locations, increasing differences in environmental conditions (e.g., temperature, rainfall, vegetation), etc. It is intuitive to use a metric proportional to the inverse distance for strengthening the predictive ability measures of $PIC$s. Let, $d_k$ be the geodesic distance\cite{karney2013algorithms} (shortest path on the surface of the earth, assuming earth to be an ellipsoid) between the target $IC$ and $PIC_k$. The normalized $[0,1]$ distance of a $PIC_k$ in the region is,

\begin{equation}
    \hat{d_k} = \frac{d_k - \min_k d_k}{\max_k d_k - \min_k d_k}
    \label{eqn:norm_distance}
\end{equation}

We want the metric to be inversely proportional to the distance. Hence, the distance metric of $PIC_k$ is defined as,

\begin{equation}
    \gamma_D(k) = 1 - \hat{d_k}
    \label{eqn:dist_weight}
\end{equation}

The outbreak history of a location is an important criterion that indicates the viral pathogen and endemic scenarios' persistence. For a $PIC_k$, we compute the prevalence $I_k$ by taking a sum of the incidence data $i_k(t)$ for the entire timeline ($\forall t \in [0, T]$).

\begin{equation}
    I_k = \sum_{t=0}^{T} i_k (t)
    \label{eqn:prev_tot}
\end{equation}

The prevalence metric is normalized $[0,1]$ across the region.

\begin{equation}
    \gamma_P(k) = \frac{I_k - \min_k I_k}{\max_k I_k - \min_k I_k}
    \label{eqn:prev_weight}
\end{equation}

\section{Results}
We present here the results in several stages. The outcomes of the feature analysis are presented first. This is followed by an analysis of prediction performance using the proposed methods. The results are generated using municipality-wise weekly dengue case data between 2010-2019 from Brazil's Espírito Santo state. We obtained data for 78 municipalities of Espírito Santo and ranked them based on the total number of cases recorded for the entire time period. The top 5 municipalities based on prevalence are listed in Table \ref{table:outbreak_sizes}. The location IDs shown in the table are the IBGE (Instituto Brasileiro de Geografia e Estatística) codes for Brazil \cite{ibge}.

\begin{table}[ht]
\caption{Top 5 locations of Espírito Santo ranked by total reported cases of dengue during 2010-2019 \cite{infoDengue}.}
\label{table:outbreak_sizes}
\centering
\begin{tabular}{|l|l|l|r|}
\hline
Loc. ID & Name & Cases & Cases per 100k \\
\hline\hline
3205309 & Vitória & 71,348 & 19,501.72 \\
3205002 & Serra & 58,424 & 11,081.10 \\
3201209 & Cachoeiro de Itapemirim & 45,319 & 21,520.12 \\
3205200 & Vila Velha & 36,743 & 7,329.18 \\
3201308 & Cariacica & 27,103 & 7,059.60 \\

\hline
\end{tabular}
\end{table}

\subsection{Feature selection and analysis}
\label{sec:feat_analysis}
\begin{figure}[ht]
\centering
\includegraphics[width=\linewidth]{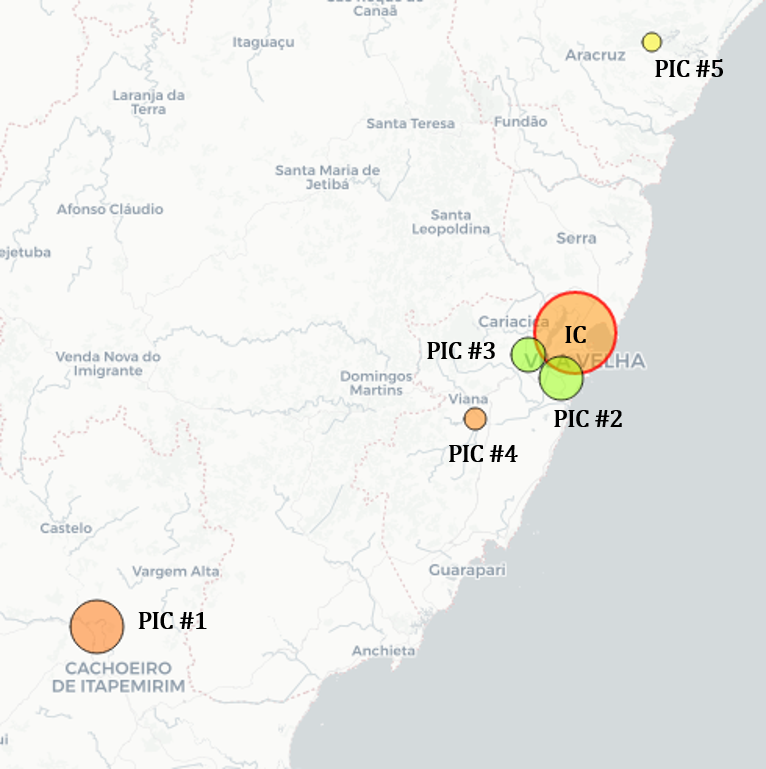}
\caption{A geospatial map showing an incidence center ($IC$) and 5 ranked peripheral incidence centers ($PIC$). The circles' radii are proportional to the total number of cases reported during 2010-2019 \cite{infoDengue}. The shades of the fill color are generated from a color gradient (green-yellow-red) which is proportional to the fraction of cases with respect to the local population of each location ($IC$ or $PIC$) during 2010-2019 \cite{infoDengue}. The greenish shades indicate smaller infected fractions, while the reddish shades indicate larger fractions. Map generated using Folium \cite{folium} with OpenStreetMap \cite{openStreetMap}. Basemap tiles provided by CartoDB \cite{cartoDB}.}
\label{fig:map_ic_pic}
\end{figure}

The $PIC$s are sorted and ranked for each $IC$, based on the predictor strength metric, $\Gamma$ (Equation \ref{eqn:pred_strength}), which is computed from the three individual metrics: correlation weight ($\gamma_C$), prevalence weight ($\gamma_P$), and distance weight ($\gamma_D$). We choose the municipality of Vitória in Espirito Santo, Brazil, as the target $IC$, which had the highest total number of cases in the state during 2010-2019, to generate the results. We compute the correlation weight using 20 fixed-length windows ($M^f=M=20$) for the time range 2010-2019, making each window approximately 26 weeks (6 months) long. The top 5 ranked $PIC$s based on $\Gamma$ are listed in Table \ref{table:pic_metrics} with the corresponding weights. While our method prioritizes the correlation weight more than others, $PIC$s with relatively lower correlation weight can still be favored because of the following factors: i) having a significant number of cases or ii) being in proximity of the target $IC$. This is evident in Table \ref{table:pic_metrics} as 3201209 ($PIC$ \#1) is chosen over 3205200 ($PIC$ \#2) and 3205101 ($PIC$ \#4) is chosen over 3200607 ($PIC$ \#5). \textit{Note, the numbers (\#) used in '$PIC$ \#' indicate rank. This should not be confused with arbitrary indices (k) used to compute metrics ($PIC_k$)}. This effect is also illustrated in Figure \ref{fig:map_ic_pic}, which shows the ranked $PIC$s listed in Table \ref{table:pic_metrics}. Although $PIC$ \#1 lies farthest from the $IC$ among the five (lowest $\gamma_D$), it is ranked at the top due to significantly higher values in the other two factors ($\gamma_C$ and $\gamma_P$).

\begin{table}[ht]
\caption{Top 5 predictor $PIC$s for Vitória. The weights based on our defined predictability metrics (correlation, prevalence, and distance) are shown in columns 3-5. The combined weights ($\Gamma$) are shown in the last column.}
\label{table:pic_metrics}
\centering
\begin{tabular}{|l|l|l|l|l|l|}
\hline
Rank \# & Loc. ID & Corr. $\gamma_C$ & Prev. $\gamma_P$ & Dist. $\gamma_D$ & $\Gamma$ \\
\hline\hline
1 & 3201209 & 0.912 & 0.812 & 0.614 & 1.301 \\
2 & 3205200 & 1.000 & 0.270 & 1.000 & 1.270 \\
3 & 3201308 & 0.956 & 0.260 & 0.996 & 1.202 \\
4 & 3205101 & 0.621 & 0.770 & 0.937 & 1.060 \\
5 & 3200607 & 0.772 & 0.524 & 0.809 & 1.029 \\
\hline
\end{tabular}
\end{table}

\begin{figure}[ht]
\centering
\includegraphics[width=\linewidth]{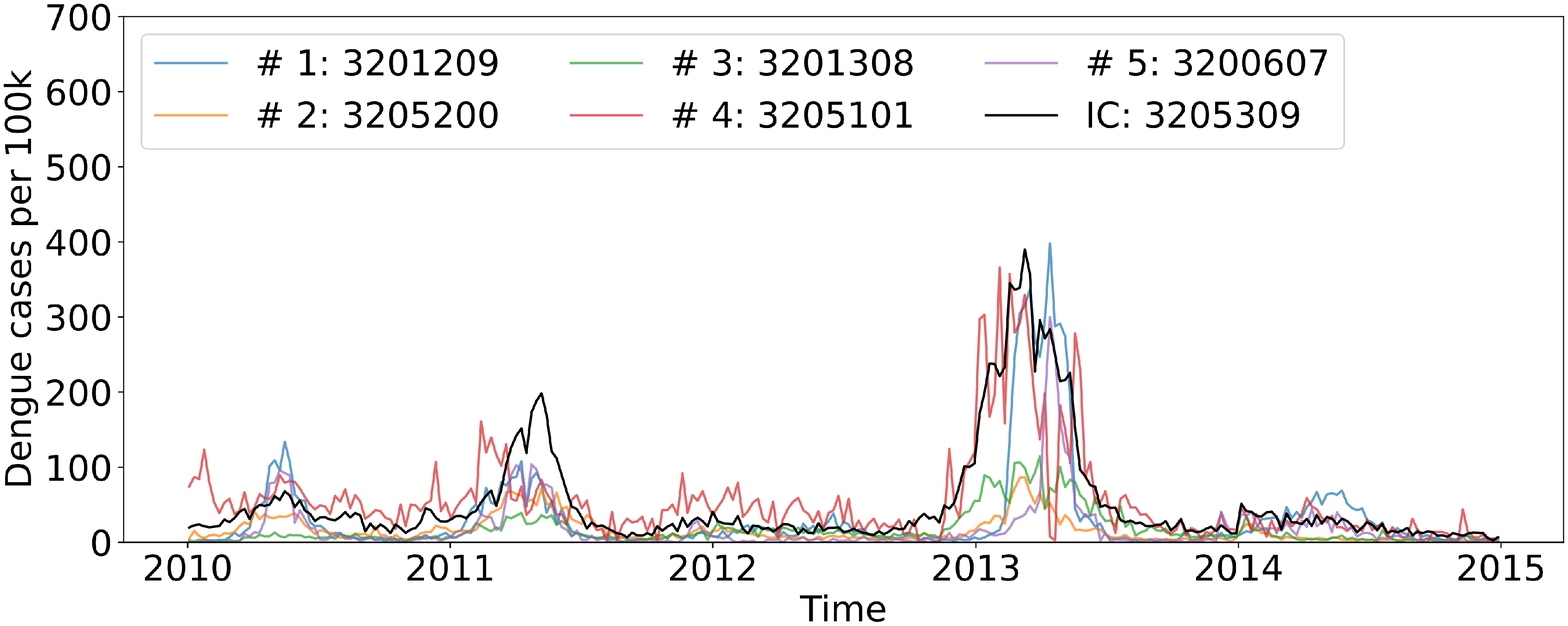}
\caption{Time series plots of weekly dengue cases per 100,000 people for the $IC$ and top 5 ranked $PIC$s (Table \ref{table:pic_metrics}). The plots depict cases only between 2010-2015 for improved clarity, but the metrics were computed based on the entire series (2010-2019).}
\label{fig:series_ic_pic}
\end{figure}

A time series plot in Figure \ref{fig:series_ic_pic} shows that the top $PIC$s are mostly correlated with the $IC$, Vitória. Among the $PIC$s displayed here, PIC \#4 (3205101) shows the weakest correlation with the $IC$. It will be clear in the upcoming results, the proximity and the high incidence fraction of this location help with prediction performance. The variability of the incidence curves prevents our method from classifying a single $PIC$ as the optimal predictor for all time ranges. However, the combination of top-ranked $PIC$s will improve prediction accuracy.

\subsection{Prediction performance}
After selecting features with the proposed methods, we train and evaluate the prediction performances using the three models (Linear, LSTM, and GRU) described in section \ref{sec:model_spec}. The time series data are split into 3 distinct sets with ratios of 50:30:20 for model training, evaluation, and testing. A sliding window with input length ($t_{in}$) of 8 and output length ($t_{out}$) 4 is used for a single shot prediction of the next 4 weeks using data of the past 8 weeks every step.

\subsubsection{Individual $IC$ prediction}
\begin{figure}[ht]
\centering
\includegraphics[width=\linewidth]{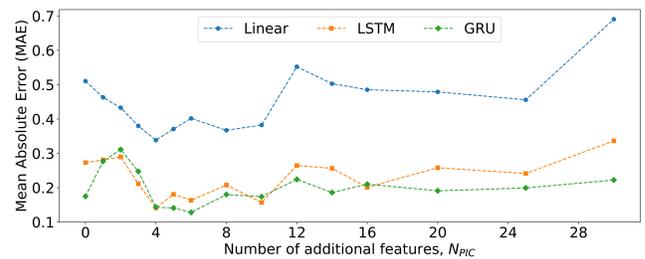}
\caption{Prediction performances of the Linear, LSTM, and GRU models in predicting normalized test data for varying number of features ($N_{PIC}$). $PIC$ data are added to the feature set based on ranks dictated by computed predictor strengths (Table \ref{table:pic_metrics}), after which models are trained and evaluated over the test data to compute the mean absolute error (MAE) metrics (lower is better).}
\label{fig:mae_vs_feat}
\end{figure}

For the $IC$ of Vitória, the $PIC$s are added gradually according to their computed ranks (section \ref{sec:feat_analysis}), and the models' prediction performances are evaluated. The mean absolute error (MAE) values on the normalized test data are plotted in Figure \ref{fig:mae_vs_feat} for varying number of additional features, $N_{PIC}$. For both LSTM and GRU models, the addition of the first two PICs deteriorates the model performance. However, the subsequent additions keep improving the outcomes. The plots quickly reach their minima, after which errors increase. The first few additions increase error due to high variability in the incidence data, as evident in Figure \ref{fig:series_ic_pic}. Further additions of $PIC$ create averaging effects on the predictor data set and cause performance improvements. The GRU model eventually reaches a minimum MAE value of 0.128, which is lower than the best optimal LSTM prediction (0.1415) by about 9.54\%. The Linear model reaches an optimum MAE value of 0.3384, which is nowhere close to the recurrent models. After reaching the minima, all three error curves rise again. This increase in MAE with larger feature sets ($N_{PIC}$) can be attributed to the curse of dimensionality \cite{hastie2009elements,li2017feature}. LSTM and GRU models perform optimally with 4 and 6 additional $PIC$s, respectively. Predicting based on present input and historical context (internal states of LSTM and GRU) of data certainly puts recurrent models ahead in performance, which is evident even without a $PIC$ ($N_{PIC} = 0$ in Figure \ref{fig:mae_vs_feat}) in the feature set. However, the linear model significantly benefits from feature addition as case data from $PIC$s strengthen inductive bias.

\begin{figure}[ht]
\centering
\includegraphics[width=\linewidth]{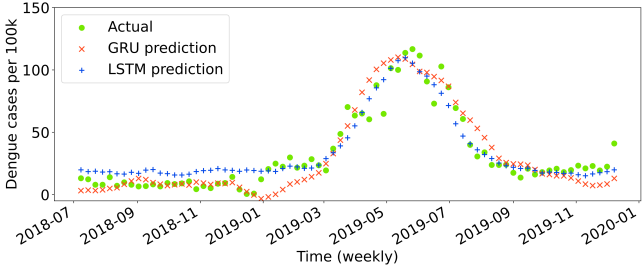}
\caption{Predicted weekly dengue cases per 100,000 people using test data with the recurrent models (LSTM and GRU). These results were obtained for the $IC$, Vitória, with 4 and 6 additional $PIC$ data joined with LSTM and GRU's feature set, respectively.}
\label{fig:series_single_ic_pred}
\end{figure}

After determining the optimum number of features ($N_{PIC}$) to be added to the predictor data set for an $IC$, recurrent models are trained with the extended feature set and are used to predict dengue cases for the test data subset of the time series. Figure \ref{fig:series_single_ic_pred} compares the predictions with actual incidence data for both LSTM and GRU models using Vitória as the $IC$ and choosing the optimum number of $PIC$s for the two models ($N_{PIC}$ = 4 for LSTM and 6 for GRU).

\subsubsection{Effect of window selection methods}
The impacts of various correlation window configurations are analyzed under the two proposed windowing methods: i) fixed-length window assignment and ii) variable-length window detection. For the first method (fixed), we vary the number of windows ($M^f$) between 5, 10, 20, and 40. For the second method (detection), we vary the minimum window size ($\Delta_{MIN}$) between 5, 10, 20, and 30 weeks. In both methods, we are consequently varying the number of windows, window lengths, and where the windows are located. In total, we run tests under 8 different scenarios and compare the prediction performances using the 3 models (Linear, LSTM, and GRU). Instead of working with a single $IC$, we run the tests over multiple locations and report the average performance metrics (e.g., mean of MAE). We take the top 20 $IC$s based on total cases per 100,000 people and average the performance metrics across $IC$s. Among the 20 $IC$s, there were 4 $IC$s where none of the models could predict with reasonable accuracy (MAE $< 1$) with or without additional features. In those locations, either the data were too limited or the outbreaks were too random for our models to generalize beyond training data. We filter out these 4 locations and take the remaining 16 locations to evaluate our methods.

\begin{figure}[ht]
\centering
\includegraphics[width=\linewidth]{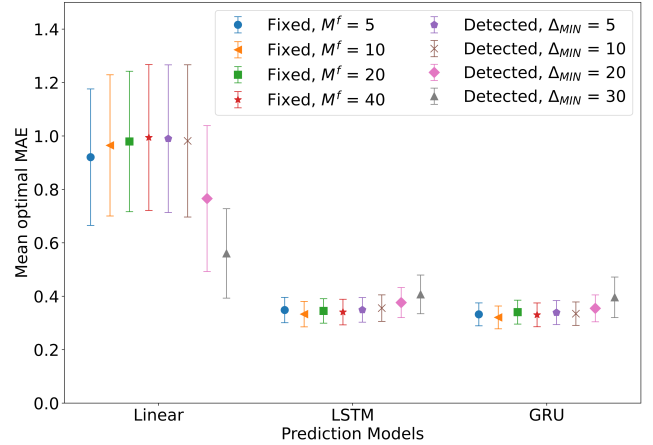}
\caption{A comparison chart of the lowest prediction error obtainable using different fixed and detected windowing schemes. Four fixed windowing schemes with the number of windows, $M^f$ = 5, 10, 20, and 40, along with four window detection schemes having minimum window lengths, $\Delta_{MIN}$ = 5, 10, 20, and 30 weeks are compared for three models. The vertical axis denotes the average of the optimal MAE values. The markers denote the mean values, with the bars representing standard errors of the mean. The results are the averages of 16 top $IC$s based on prevalence.}
\label{fig:win_compare}
\end{figure}

For a model, predicting on a given $IC$, the optimal MAE is defined as the minimum mean absolute error (MAE) obtained by varying the number of additional features ($N_{PIC}$) from the $PIC$ ranked list produced by a given windowing method (i.e., minima of the curves in Figure \ref{fig:mae_vs_feat} are the optimal MAEs of three models). The average optimal MAE (across 16 $IC$s) are plotted in Figure \ref{fig:win_compare} for all 8 windowing schemes. The recurrent models perform significantly better than the Linear model: on average, LSTM and GRU outperform the Linear model by 60\% and 61\%, respectively. LSTM and GRU outperform the best-performing linear model by 41\% and 43\%, respectively. We expect this kind of advantage, given the importance of considering historical data in predicting future dengue cases. The performances across different windowing schemes with LSTM and GRU models are comparable (both models have an approximate standard deviation of 0.02 around their means of 0.3569 and 0.3435, respectively), with a small trend of increasing mean error values with windowing method configurations. In the case of linear models, there are some stark contrasts when using window detection methods. Using the window detection method with a large $\Delta_{MIN}$ value increases performance sharply. For example, using a $\Delta_{MIN}$ of 30 weeks shows 41.9\% improvement over fixed window schemes' average performance. This indicates that features selected with a small number of large correlation windows detected based on outbreak locations in the time series are more effective than the remaining schemes that use a relatively large number of smaller windows. A similar trend is also visible with fixed window methods, although it is not statistically conclusive. The lower values of $M^f$, which translates to having a lower number of large windows, show slightly better performance over higher values of $M^f$.

\begin{figure}[ht]
\centering
\includegraphics[width=\linewidth]{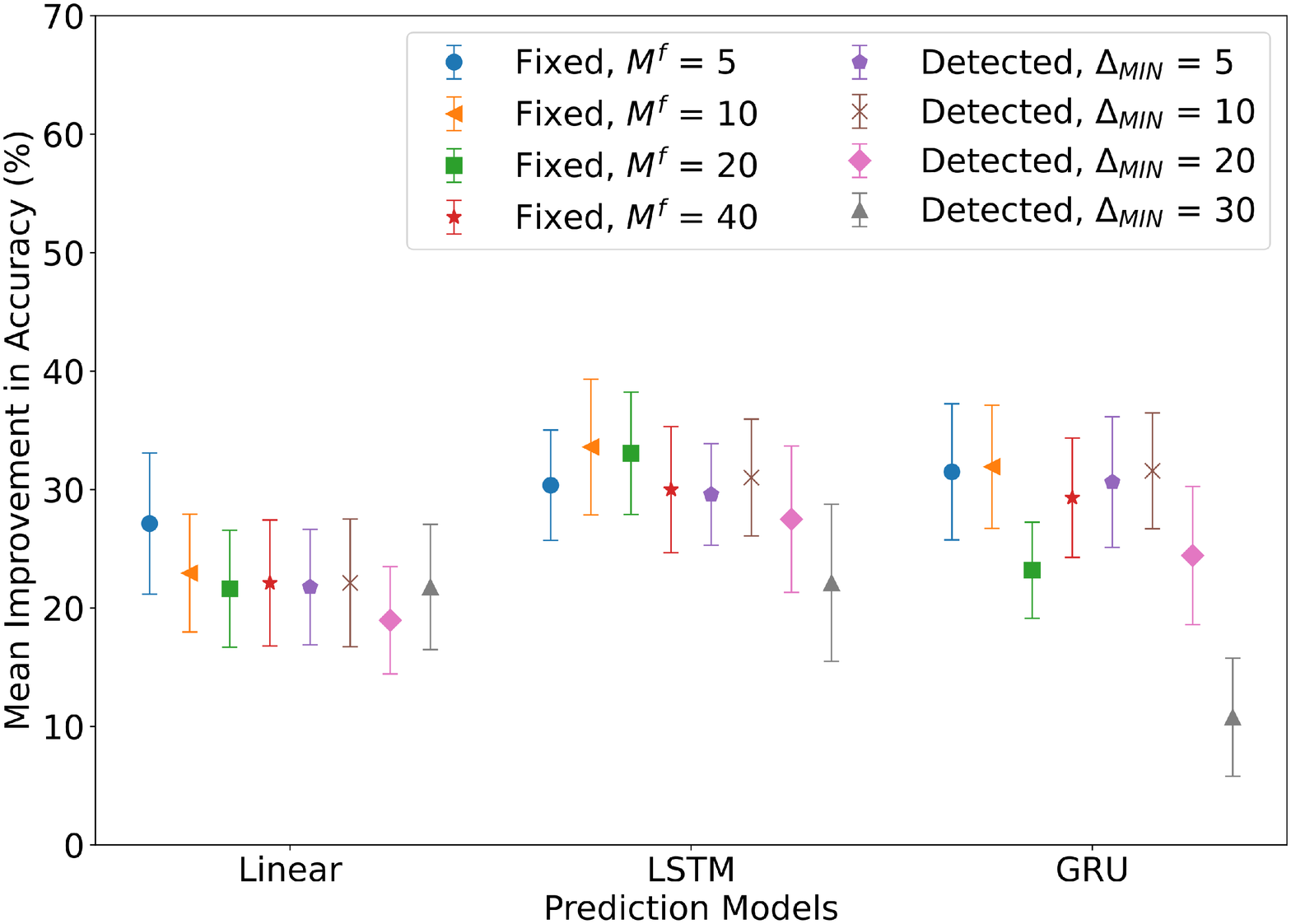}
\caption{A comparison chart of average improvement in accuracy (reduction in MAE) with respect to performance without feature enhancements ($N_{PIC}$ = 0) for different fixed and detected windowing schemes. Four fixed windowing schemes with the number of windows, $M^f$ = 5, 10, 20, and 40, along with four window detection schemes having minimum window lengths, $\Delta_{MIN}$ = 5, 10, 20, and 30 weeks are compared for three models. The markers denote the mean values, with the bars representing standard errors of the mean. The results are the averages of 16 top $IC$s based on prevalence.}
\label{fig:win_improvements}
\end{figure}

To understand how beneficial our proposed methods are over the baseline (without feature enhancement, $N_{PIC}$ = 0), we analyze the improvements in prediction accuracy. The term, \textit{improvement in accuracy} is defined as the relative decrease in MAE (i.e., increase in prediction performance) with optimum $PIC$ selection (Figure \ref{fig:win_compare}) compared to MAE without additional features ($N_{PIC} = 0$). We compute the accuracy improvements across all 16 $IC$s and plot the \textit{mean improvement in accuracy} as percentages in Figure \ref{fig:win_improvements}. The linear model demonstrates average improvements ranging from 18.97\% to 27.13\% depending on the window selection schemes. The LSTM model improvements range from 22.13\% to 33.6\% and the GRU model range from 10.79\% to 31.92\% depending on the window selection schemes. We should be careful in interpreting these values; greater improvements do not translate to better optimal performance. These values are relative to their baselines ($N_{PIC} = 0$). With some incidence centers ($IC$), a model can already predict with higher accuracy than other incidence centers. Our methods help minimize those gaps and still provide some improvements over the baselines (ranging from 10.79\% to 33.6\% depending on the model and the scheme). Figure \ref{fig:win_compare} demonstrates that, in general, GRU and LSTM both perform well when we deal with average values. For individual $IC$s, however, a conclusive determination of the best performing model can be done (either GRU or LSTM).

\subsubsection{Performance on aggregated data}
It is sometimes reasonable to aggregate data to larger scales based on geographic adjacency and environmental similarities (e.g., weather). From a macroscopic point of view, predicting for an ecoregion may be more meaningful for policymakers to interpret the outcomes. According to Omernik (2004), ecoregions are defined as areas within which there is a spatial coincidence in characteristics of geographical phenomena (e.g., geology, physiography, vegetation, land use, climate, hydrology, terrestrial and aquatic fauna, etc.) associated with differences in the quality, health, and integrity of ecosystems \cite{omernik2004perspectives}. We use the terrestrial ecoregions defined by The Nature Conservancy(TNC) in this work \cite{olson2002global}. The following analysis is performed for Brazilian locations with available data in the Bahia Coastal Forest ecoregion.

\begin{figure}[ht]
\centering
\includegraphics[width=\linewidth]{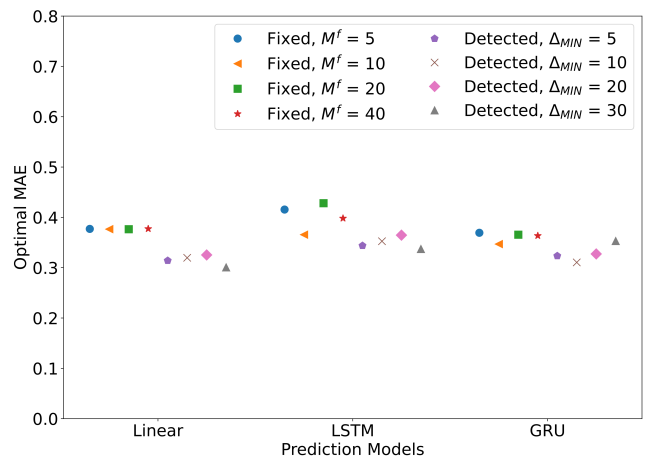}
\caption{A comparison chart of lowest prediction error obtainable using different fixed and detected windowing schemes for case data aggregated across an ecoregion. Four fixed windowing schemes with the number of windows, $M^f$ = 5, 10, 20, and 40, along with four window detection schemes having minimum window lengths, $\Delta_{MIN}$ = 5, 10, 20, and 30 weeks are compared for three models. The vertical axis denotes the optimum mean absolute error (MAE). The results were obtained for the ecoregion named Bahia Coastal Forest \cite{olson2002global}.}
\label{fig:win_compare_ecoregion}
\end{figure}

Comparing the prediction performance on aggregated data shows that the advantages of recurrent models compared to the Linear model (which we had for individual $IC$s) are diminished. The optimal performances across prediction models become more comparable, as shown in Figure \ref{fig:win_compare_ecoregion}. The mean optimal MAE values obtained for 4 fixed window schemes are 0.38, 0.40, and 0.36 when using Linear, LSTM, and GRU models, respectively. The mean optimal MAEs for 4 window detection schemes are 0.31, 0.35, and 0.33 using the same three models. We observe a distinct advantage of the window detection method for selecting $PIC$s. On average, the window detection methods improve over their baselines ($N_{PIC}=0$) by 16.50\%, 12.68\%, and 10.07\% for Linear, LSTM, and GRU models, respectively. Variable window allocation based on outbreak window detection outperforms fixed window allocation methods. In other words, windowed cross-correlation on a few important outbreak regions performs better than comparing over the entire time series. As the target data is aggregated from all $PIC$s in the region, the optimally trained Linear models sometimes perform better than recurrent models. One key takeaway is that cases for the entire region can be predicted with improved accuracy while using a small subset ($PIC$s) of data as features. For the results shown in Figure \ref{fig:win_compare_ecoregion}, the optimal MAE values can be reached for $N_{PIC}$ values between 2 and 5. This aggregation provides a fast way to predict cases, with a small fraction of regional outbreak data. While the accuracy achieved at the ecoregion level is not on par with the accuracy achieved for single $IC$s, this prediction is useful to generate risk maps.

\begin{figure}[ht]
\centering
\includegraphics[width=0.9\linewidth]{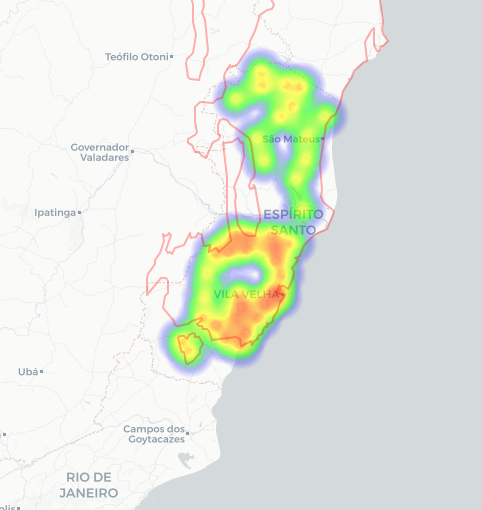}
\caption{A geospatial map showing the predicted risk of infection in an ecoregion of Brazil. A part of the ecoregion (Bahia Coastal Forests) shown here is marked with red borders. The risk of infection is shown as a heatmap with color shades ranging from blue (low risk) to red (high risk). The heatmap is constructed by combining the ranked $PIC$s in the ecoregion, with higher-ranked $PIC$s contributing more to the risk. The aggregated data was predicted using the Linear model with an optimum number of features ($N_{PIC} = 4$) selected with window detection method ($\Delta_{MIN}=30$). This heatmap depicts the risk on the date of 09-June-2019, with 40 top-ranked $PIC$. Map generated using Folium \cite{folium} with OpenStreetMap \cite{openStreetMap}. Basemap tiles provided by CartoDB \cite{cartoDB}.}
\label{fig:risk_ecoregion}
\end{figure}

We construct a risk map for the Bahia Coastal Forest ecoregion using the predicted cases and weights of the top-ranked $PIC$ in Figure \ref{fig:risk_ecoregion}. The top 40 $PIC$s are included in the map, and the $PIC$s that contribute more towards the weight ($\Gamma$) are shown as high-risk regions (e.g., red). The aggregated prediction was done using the optimal Linear model with $\Delta_{MIN}=30$ and $N_{PIC} = 4$.

\section{Conclusion}
In this work, we develop a method to select relevant incidence data from peripheral locations as features to improve the prediction of dengue fever outbreaks. In order to rank features, we use windowed cross-correlation analysis on dengue case data. We propose two methods for allocating correlation windows (position and size) over the time series to compute correlation weights. For a target location ($IC$), peripheral locations ($PIC$) are ranked based on a combination of correlation, distance, prevalence metrics. The predictive models benefit from the ranked feature sets, as these reach model and location-specific optimal performances with a relatively small subset of features. We tested three predictive models using dengue case data from Brazil, showing different levels of accuracy gains.

On average, the proposed feature enhancement methods improve prediction performance by 10.79\% to 33.6\% over the baseline feature set for the locations we tested, depending on the prediction model and the window allocation scheme. For the location with the highest total cases (2010-2019) in the Espírito Santo region of Brazil, we could get MAE values as low as 0.13 (normalized case data) using the GRU model with data from just 6 locations added to the feature set. In a test across multiple locations, both RNN models (LSTM and GRU) performed with comparable accuracy (average MAE ranging from 0.3435 to 0.3569) when using an optimal number of additional features. The Linear model also benefited (18.97\% to 27.13\% improvement over the baseline) from windowed correlation-based feature enhancements, although its performance never got close to recurrent models. When compared with the respective optimal number of features, the best performing recurrent models outperformed the Linear model by at least 41\% in terms of prediction accuracy. The window detection methods showed performance comparable to fixed window allocation. This can be advantageous when working with extensive sets of data, as the detected windows only compare a subset of important time steps instead of the entire series. For municipality-level dengue case prediction, GRU was the best performing model, closely followed by LSTM. The performance gaps between these two models diminished after feature set optimization. When predicting aggregated data for the entire region using a subset of constituent locations, our methods reach optimal performance with the addition of only 2-5 locations (out of 77), depending on the model and window selection scheme. This is especially useful for situations where the lack of trainable data hinders forecasting. For example, dengue risk for a region can be predicted if a small subset of data from epidemically important region locations is available. This addresses the issue of incomplete data and eases the `curse of dimensionality` while improving training efficiency.

Future efforts in this area can focus on gaining further insights based on epidemiological, environmental, and economic characteristics of locations and combining them to improve feature ranks. While case data are indicators of the severity of an outbreak, these are not always adequate to explain future possibilities. Machine learning models cannot generalize beyond training data for every location with the same degree of accuracy. Complex interactions of dengue viral strains and changes in host immunity patterns may evolve outbreak characteristics over the years. Several random factors, including host travel patterns, natural disasters, and lifestyle changes because of other infectious outbreaks (e.g., COVID-19 pandemic), may affect dengue outbreaks. While the metrics used in our method capture such factors' long-term characteristics, these do not account for randomness. While keeping the feature sets reasonably small, our method can improve outbreak prediction. The proposed method can be generalized and used for projecting any infectious outbreak where temporal data at a reasonable spatial resolution are available.

\section*{Acknowledgment}
This research has been supported by the Department of the Army, U.S. Army Contracting Command, Aberdeen Proving Ground, Natick Contracting Division, Fort Detrick (DWFP Grant W911QY-19-1-0004).


\ifCLASSOPTIONcaptionsoff
  \newpage
\fi



\bibliographystyle{IEEEtran}
\end{document}